\documentclass[conference,letterpaper]{IEEEtran}
\IEEEoverridecommandlockouts

\ifCLASSINFOpdf
\else
\usepackage[dvips]{graphicx}
\fi
\usepackage{url}
\usepackage{graphicx} 
\usepackage{booktabs} 
\usepackage{mathrsfs}
\usepackage{amsmath}
\usepackage{amssymb}
\usepackage{mathtools}
\usepackage{cite}
\usepackage[acronym,shortcuts]{glossaries}
\usepackage{lipsum}
\usepackage{xcolor}

\usepackage{comment}

\usepackage{algorithm, algpseudocode}

\usepackage{bm}
\usepackage{mleftright}
\usepackage{orcidlink}
\usepackage{fix-cm}
\usepackage{pifont}
\DeclareMathOperator*{\argmin}{arg\,min}

\newacronym{GA}{GA}{genie-aided}
\newacronym{EA}{EA}{``\emph{estimate-then-average}''}
\newacronym{AE}{AE}{``\emph{average-then-estimate}''}
\newacronym{IRS}{IRS}{intelligent reflecting surface}
\newacronym{RSSI}{RSSI}{received signal strength indicator}
\newacronym{SotA}{SotA}{state-of-the-art}
\newacronym{CSI}{CSI}{channel state information}
\newacronym{D2D}{D2D}{device-to-device}
\newacronym{RR}{RR}{round-robin}
\newacronym{DA}{DA}{Dutch auction}
\newacronym{CWFL}{CWFL}{clustered WFL}
\newacronym{WFL}{WFL}{wireless federated learning}
\newacronym{RSMA}{RSMA}{rate splitting multiple access}
\newacronym{IoT}{IoT}{Internet-of-Things}
\newacronym{TDMA}{TDMA}{time-domain multiple access}
\newacronym{NOMA}{NOMA}{non-orthogonal multiple access}
\newacronym{ML}{ML}{machine learning}
\newacronym{MIMO}{MIMO}{multiple-input multiple-output}
\newacronym{CT}{CT}{compute-then-transmit}
\newacronym{FP}{FP}{fractional programming}
\newacronym{CF-mMIMO}{CF-mMIMO}{cell free massive MIMO}
\newacronym{iid}{i.i.d.}{independent and identically distributed}
\newacronym{DL}{DL}{downlink}
\newacronym{UL}{UL}{uplink}
\newacronym{IC}{IC}{interference cancellation}
\newacronym{SIC}{SIC}{successive interference cancellation}
\newacronym{SI}{SI}{soft-impute}
\newacronym{BS}{BS}{base station}
\newacronym{TX}{TX}{transmit}
\newacronym{RX}{RX}{receive}
\newacronym{MU}{MU}{multi-user}
\newacronym{SISO}{SISO}{single-input single-output}
\newacronym{AWGN}{AWGN}{additive white Gaussian noise}
\newacronym{SINR}{SINR}{signal-to-interference-and-noise ratio}
\newacronym{FL}{FL}{federated learning}
\newacronym{CPU}{CPU}{central processing unit}
\newacronym{KNN}{KNN}{K-nearest-neighbor}
\newacronym{RF}{RF}{radio frequency}
\newacronym{GD}{GD}{gradient descent}

\newacronym{RSS}{RSS}{received signal strength}
\newacronym{FIM}{FIM}{fisher information matrix}
\newacronym{ToA}{ToA}{time of arrival}
\newacronym{AoA}{AoA}{angle of arrival}
\newacronym{GP}{GP}{Gaussian process}
\newacronym{2D}{2D}{two-dimensional}
\newacronym{GPR}{GPR}{Gaussian process regression}
\newacronym{GNSS}{GNSS}{global navigation satellite systems}
\newacronym{B5G}{B5G}{beyond fifth-generation}
\newacronym{RRH}{RRH}{remote radio head}
\newacronym{GPS}{GPS}{Global Positioning System}
\newacronym{RFID}{RFID}{radio frequency identification}
\newacronym{TCAS}{TCAS}{traffic alert and collision avoidance systems}
\newacronym{RMSE}{RMSE}{root mean square error}
\newacronym{SGD}{SGD}{stochastic gradient descent}
\newacronym{PDF}{PDF}{probability density function}
\newacronym{CU}{CU}{computing unit}
\newacronym{DM-MIMO}{DM-MIMO}{distributed massive multiple-input multiple-output}
\newacronym{LOS}{LOS}{line-of-sight}
\newacronym{NLOS}{NLOS}{non-line-of-sight}
\newacronym{ROI}{ROI}{region of interest}
\newacronym{AP}{AP}{access point}
\newacronym{TDOA}{TDOA}{time difference of arrival}
\newacronym{UE}{UE}{user equipment}
\newacronym{dB}{dB}{decibel}
\newacronym{RIS}{RIS}{reconfigurable intelligent surface}
\newacronym{CG}{CG}{conjugate gradient}

\newacronym{PG}{PG}{proximal gradient}
\newacronym{SVT}{SVT}{singular value thresholding}
\newacronym{NN}{NN}{nuclear norm}
\newacronym{NMSE}{NMSE}{normalized mean square error}
\newacronym{MC}{MC}{matrix completion}
\newacronym{NP}{NP}{non-deterministic polynomial-time}
\newacronym{SDP}{SDP}{semidefinite programming}

\newacronym{TC}{TC}{tensor completion}
\newacronym{ADMM}{ADMM}{alternating direction method of multipliers}

\def\BibTeX{{\rm B\kern-.05em{\sc i\kern-.025em b}\kern-.08em
    T\kern-.1667em\lower.7ex\hbox{E}\kern-.125emX}}
\begin{document}

\title{Low Rank Tensor Completion via Adaptive ADMM}

\author{

\IEEEauthorblockN{Niclas F\"uhrling\textsuperscript{\orcidlink{0000-0003-1942-8691}}, Getuar Rexhepi\textsuperscript{\orcidlink{0009-0002-3268-522X}}, and Giuseppe Thadeu Freitas de Abreu\textsuperscript{\orcidlink{0000-0002-5018-8174}}}
\IEEEauthorblockA{\textit{School of Computer Science and Engineering}, 
\textit{Constructor University},
Bremen, Germany \\
\{nfuehrling,grexhepi,gabreu\}@constructor.university}
\vspace{-5ex}
}

\maketitle
\begin{abstract}
We consider a novel algorithm, for the completion of partially observed low-rank tensors, as a generalization of matrix completion.
The proposed low-rank \ac{TC} method builds on the conventional \ac{NN} minimization-based low-rank \ac{TC} paradigm, by leveraging the \ac{ADMM} optimization framework.
To that extend the original \ac{NN} minimization problem is reformulated into multiple subproblems, which are then solved iteratively via closed-form proximal operators, making use of over-relaxation and an adaptive penalty parameter update scheme, to further speed up convergence and improve the overall performance of the method.
Simulation results demonstrate the superior performance of the new method in terms of \ac{NMSE}, compared to the conventional \ac{SotA} techniques, including \ac{NN} minimization approaches, as well as a mixture of the latter with a matrix factorization approach, while its convergence can be significantly improved by initializing the algorithm with the solution of the \ac{SotA}.

\end{abstract}


\glsresetall
\IEEEpeerreviewmaketitle

\section{Introduction}

\IEEEPARstart{D}{ue} to the rise of machine learning and big data, \ac{MC} has emerged as a fundamental problem in a wide range of modern applications, that span from localization algorithms in signal processing \cite{Nguyen_2019_Loc}, to collaborative filtering and recommendation systems \cite{Chen_2022}, to more recent applications such as millimeter-wave channel estimation in wireless communications \cite{Vlachos_2018}, and many more.

Conventional \ac{MC} techniques are based on a structured low-rank problem formulation \cite{Candes_2009_Noise,Candes_2009,Candes_2010}, with the goal a low-rank matrix $\boldsymbol{X}\in\mathbb{R}^{m\times n}$ from a partially observed and incomplete matrix $\boldsymbol{O}\in\mathbb{R}^{m\times n}$ \cite{Nguyen_2019 , Dai_2012 , Bart_2013}.

However, the increasing use of multidimensional data in various domains has highlighted the limitations of matrix completion, which has motivated the extension to \acf{TC}.

Similar to matrix completion, tensor completion aims to recover a low-rank tensor $\mathcal{\boldsymbol{T}} \in \mathbb{R}^{I_1 \times I_2 \times \cdots \times I_N},$ from a partially observed tensor $\mathcal{\boldsymbol{O}}$ of the same dimensions, under the assumption that the true tensor admits a low-dimensional latent structure \cite{Song_2019,Ji_2019,Wang_2025}. Unlike matrices, however, tensors admit multiple notions of rank, such as CP rank, Tucker rank, and tubal rank, each capturing different types of multilinear dependencies. 
This fundamental difference makes \ac{TC} both more expressive and significantly more challenging than its matrix counterpart.

Early \ac{TC} approaches have largely been inspired by developments in \ac{MC}, where the non-convex tensor rank objectives are replaced with convex or tractable surrogates.
Common techniques rely on minimizing sums of nuclear norms of mode-n unfoldings, effectively extending \ac{NN}-based \ac{MC} techniques to the tensor setting, as proposed in \cite{Liu_2013}. 
While such relaxations preserve convexity and theoretical guarantees, they often suffer from high computational complexity and may fail to fully exploit the intrinsic multilinear structure of the tensor. 
As an alternative, factorization-based formulations, which parameterize the tensor through low-dimensional latent factors, have been widely adopted due to their favorable scalability and reduced memory footprint \cite{xu2013parallel}.

In addition, hybrid methods that combine nuclear-norm regularization with factorized representations have been proposed to bridge the gap between convex relaxations and scalable low-rank modeling \cite{Gao2018}. 
While more recent advances have proposed discrete-aware techniques for both matrix and tensor completion \cite{Iimori_2020,Nic_Asilo_2024,Nic_DaLRTC}, these methods, while having shown superior performance compared to their continuous counterparts, are only applicable in limited applications.

In light of the above, we propose an extension of the \acf{TC} paradigm via a novel low-rank \ac{TC} method, that leverages \acf{ADMM} to solve the \ac{NN} minimization problem, by reformulating the original problem into multiple subproblems.
These subproblems can then be solved iteratively in closed-form, also leveraging over-relaxation and an adaptive penalty parameter update scheme.

The rest of the article is structured as follows.
First, basics on the \ac{TC} problem, as a generalization of the \ac{MC} problem, with a description of conventional \ac{TC} techniques is given in Section \ref{sec:prior}.
Then, the proposed \ac{ADMM} method is described in Section \ref{sec:proposed}, followed by numerical results that compare our contribution with \ac{SotA} methods and a few concluding remarks, in Sections \ref{sec:results} and \ref{sec:conclusions}, respectively.

\section{Preliminaries}
\label{sec:prior}

In general, tensors are a multi-dimensional generalization of vectors and matrices, such that an N-mode tensor $\mathcal{T}$ can be written as
\begin{equation}
    \mathcal{T} \in \mathbb{R}^{I_1 \times I_2 \times \cdots \times I_N},
\end{equation}
where $I_n$ denotes the size of the tensor in its $n$-th mode.

For the sake of simplicity, we perform the tensor completion in terms of image processing, where a partially observed, sampled version of an image is received, which needs to be recovered.

\subsection{Conventional Tensor Completion Problem Formulation}
Conventional \ac{SotA} low rank \ac{TC} techniques are based on a minimizing the rank of the tensor.
However, to introduce the concept, the problem can be simplified to a \ac{MC} problem, such that the original rank minimization can be written as
\vspace{-1ex}
\begin{subequations}
\label{eq:rank_opt}
\begin{align}
\argmin_{\boldsymbol{X}\in \mathbb{R}^{m\times n}}&\quad \text{rank}(\boldsymbol{X}),\\
\text{s.t. }&\quad P_{\Omega}(\boldsymbol{X})=P_{\Omega}(\boldsymbol{O}),
\vspace{-1ex}
\end{align}
\end{subequations}
where $\text{rank}(\cdot)$ denotes the rank operator of the input matrix, and $P_{\Omega}(\cdot)$ indicates the mask operator, which is defined as 
\vspace{-1ex}
\begin{equation}
[P_{\Omega}(\boldsymbol{X})]_{i,j}=
\begin{cases}
[\boldsymbol{X}]_{i,j},& \text{if } (i,j) \in \Omega, \\
0,              & \text{otherwise},
\end{cases}
\vspace{-1ex}
\end{equation}
where $\Omega$ defines the index set that contains the indices of the observed elements and $[\cdot]_{i,j}$ denotes the $(i,j)$-th element of a given matrix.

While problem \eqref{eq:rank_opt} is optimal, it has been shown in \cite{Hillar_2013} that the problem is \ac{NP}-hard due to the non-convexity of the rank operator, such that its solution cannot be obtained efficiently, which applies to both, matrices and tensors.

However, it has also been shown in \cite{Candes_2012} that the rank minimization problem can be relaxed by replacing the rank objective with the \ac{NN} operator $||\boldsymbol{X}||_*$, which is given by the sum of the singular values of $\boldsymbol{X}$, since a matrix of rank $r$ has exactly $r$ nonzero singular values.
The relaxed optimization problem utilizing the \ac{NN} objective can be therefore written as
\vspace{-3ex}
\begin{subequations}
\label{eq:NNproblem}
\begin{align}
\argmin_{\boldsymbol{X}\in \mathbb{R}^{m\times n}}&\quad ||\boldsymbol{X}||_*,\\
\text{s.t. }&\quad P_{\Omega}(\boldsymbol{X})=P_{\Omega}(\boldsymbol{O}) ,
\label{eq:C1}
\end{align}
\end{subequations}
which can be viewed as a tight lower bound of the rank operator \cite{Recht_2010}, compared to the original problem in \eqref{eq:rank_opt}.


Even though enforcing constraint \eqref{eq:C1} exactly is technically correct, in practical scenarios, the observed matrix typically contains measurement noise.
To that extent, the low-rank requirement can be relaxed, such that the target matrix is recovered as an approximately low-rank solution.
In this context, several related works, $e.g.$, \cite{OptSpace, RankEDM, Wong_2017}, have considered the following relaxed formulation of problem \eqref{eq:NNproblem}, defined as
\begin{subequations}
\label{eq:NN_opt}
\begin{align}
\argmin_{\boldsymbol{X}\in \mathbb{R}^{m\times n}}&\quad ||\boldsymbol{X}||_*,\\[-1ex]
\text{s.t. }&\quad \underbrace{\frac{1}{2}||P_{\Omega}(\boldsymbol{X}-\boldsymbol{O}) ||^2_F}_{\triangleq f(\boldsymbol{X})}\leq \epsilon,
\end{align}
\end{subequations}
which can be rewritten in regularized form as
\begin{equation}
\label{eq:NN_opt_reg}
\argmin_{\boldsymbol{X}\in \mathbb{R}^{m\times n}}\quad f(\boldsymbol{X})+\lambda||\boldsymbol{X}||_*.
\end{equation}
%

Next, as mentioned before, the same concept of nuclear norm minimization can be applied to Tensors.
For $\bm{\mathcal{X}},\bm{\mathcal{T}}=\mathbb{R}^{\bm{I}_1,\bm{I}_1,\cdots \bm{I}_N}$ being N-mode Tensors the problem can be rewritten as
\vspace{-2ex}
\begin{subequations}
\begin{align}
    \argmin_{ \bm{\mathcal{X}}} &\quad||\boldsymbol{\mathcal{X}}||_*,\\
    \text{s.t} &\quad P_{\Omega}(\boldsymbol{\mathcal{X}})=P_{\Omega}(\boldsymbol{\mathcal{T}}), 
\end{align}
\label{eq:NN_opt_og}
\end{subequations}
where the nuclear norm for a Tensor is defined as 
\begin{equation}
 ||\boldsymbol{\mathcal{X}}||_*=\sum^N_{n=1}\alpha_n||\boldsymbol{X}_{(n)}||_*
\end{equation}
where $\bm{X}_{(n)}$ denotes the mode-\(n\) matricization of the tensor $\bm{\mathcal{X}}$ and $\alpha_n\geq 0,\sum{\alpha_n}=1$, which can be further rewritten as
\begin{subequations}
\begin{align}
    \argmin_{ \bm{\mathcal{X}}}&\quad \sum^N_{n=1}\alpha_n ||\boldsymbol{X}_{(n)}||_*,\\
    \text{s.t} &\quad P_{\Omega}(\boldsymbol{\mathcal{X}})=P_{\Omega}(\boldsymbol{\mathcal{T}}).
\end{align}
\end{subequations}

To elaborate, mode-\(n\) matrizitation rearranges the elements of a tensor into a \ac{2D} matrix.
For a tensor \(\boldsymbol{\mathcal{T}} \in \mathbb{R}^{I_1 \times I_2 \times \cdots \times I_N}\), the mode-\(n\) unfolding is a matrix, defined as
\begin{equation}
    \boldsymbol{T}_{(n)} \in \mathbb{R}^{I_n \times \prod_{j \neq n} I_j}.
\end{equation}

\subsection{\Acl{SI} Approach}
A straightforward approach to solve the \ac{NN} minimization problem in \eqref{eq:NN_opt_reg} is to use the \ac{SI} method, which is not only relevant in the context of matrix completion, but also in the context of tensor completion, since the same concept can be applied to the mode-n matricization of the tensor, as discussed above.
To be more specific, \Ac{SI} \cite{Mazumder_2010} and its extension the accelerated and inexact \ac{SI} (AIS)-impute approach of \cite{Yao_2019}, are recently proposed methods that solve large-scale \ac{MC} problems \cite{Recht_2013,Fang_2017}, which have a similar structure as the problems described by equations \eqref{eq:NN_opt} and \eqref{eq:NN_opt_reg} that aim to solve the \ac{NN} minimization problem efficiently by using \ac{SVT}. 
In the conventional \ac{SI} approach the following recursion is used, given by
\begin{equation}
\label{eq:soft_inpute}
\boldsymbol{X}_t=\text{SVT}_\lambda(\boldsymbol{X}_{t-1}+P_{\Omega}(\boldsymbol{O}-\boldsymbol{X}_{t-1})),
\end{equation}
where the \ac{SVT} function is defined as
\begin{equation}
\text{SVT}_\lambda(\boldsymbol{A})=\boldsymbol{U}(\boldsymbol{\Sigma}-\lambda\boldsymbol{I})_+\boldsymbol{V}^\intercal,
\end{equation}
in which $(\cdot)_+$ defines the positive part of the input and $\boldsymbol{A}\triangleq\boldsymbol{U}\boldsymbol{\Sigma}\boldsymbol{V}^\intercal$.

\vspace{-1ex}
\section{Proposed Tensor Completion Method}
\label{sec:proposed}

\subsection{Problem Formulation}
\label{sec:formulation}

Let $\mathcal{T} \in \mathbb{R}^{I_1 \times I_2 \times I_3}$ be a partially observed low-rank tensor, with the set of observed indices denoted as $\Omega$ and the corresponding masking operator $\mathcal{P}_\Omega(\cdot)$.  
Revisiting the problem in \eqref{eq:NN_opt_og}, we seek to recover $\mathcal{T}$ by solving the Sum-of-Nuclear-Norms minimization problem, written as
\vspace{-1ex}
\begin{subequations}
    \label{eq:snn}
  \begin{align}
      \argmin_{\mathcal{X} \in \mathbb{R}^{I_1 \times I_2 \times I_3}}&\quad
    \lambda \sum_{n=1}^{N} \alpha_n \bigl\|\boldsymbol{X}_{(n)}\bigr\|_*,\\
  \text{s.t.}&\quad
  \mathcal{P}_\Omega(\mathcal{X}) = \mathcal{P}_\Omega(\mathcal{T}),
    \end{align}
\end{subequations}
where $\boldsymbol{X}_{(n)} \in \mathbb{R}^{I_n \times \prod_{j \neq n} I_j}$ is the mode-$n$ unfolding of $\mathcal{X}$, $\alpha_n \geq 0$ are mode-wise weights satisfying $\sum_n \alpha_n = 1$, and $\lambda > 0$ controls the degree of nuclear-norm regularization.

Problem \eqref{eq:snn} was originally addressed by the algorithm in \cite{Liu_2013}, which solve it via \ac{ADMM} with a fixed penalty parameter.  
While convergent, fixed-penalty \ac{ADMM} converges at a rate $O(1/k)$, and its empirical speed depends critically on the (manually chosen) penalty value.  
We propose a significantly improved solver, that combines three well-established \ac{ADMM} accelerators, namely, adaptive penalty, over-relaxation, and a principled stopping criterion in a single algorithm, that is shown to significantly outperform the \ac{SotA}.

To prepare the problem for the use of \ac{ADMM}, we introduce $N$ auxiliary matrix variables $\mathbf{M}_n \in \mathbb{R}^{I_n \times \prod_{j \neq n} I_j}$ to represent each
mode-$n$ unfolding independently.  
The problem presented in \eqref{eq:snn} can therefore be rewritten as
\begin{align}
  \argmin_{\mathcal{X},\,\{\mathbf{M}_n\}} & \quad
    \lambda \sum_{n=1}^{N} \alpha_n \|\mathbf{M}_n\|_*
  \label{eq:split_obj}\\
  \text{s.t.} &\quad
    \boldsymbol{X}_{(n)} = \mathbf{M}_n, \quad n = 1,\ldots,N,
  \label{eq:split_eq}\\
  &\quad
    \mathcal{P}_\Omega(\mathcal{X}) = \mathcal{P}_\Omega(\mathcal{T}).
  \label{eq:split_obs}
\end{align}

By the introduction of scaled dual variables $\mathbf{U}_n$ and a penalty
$\rho > 0$, the augmented Lagrangian is given by
\begin{align}
    \label{eq:AL}
  \mathcal{L}_\rho\!\left(\mathcal{X},\mathbf{M}_n,\mathbf{U}_n\right)
  &=
  \lambda \sum_{n=1}^{N} \alpha_n \|\mathbf{M}_n\|_*\\ \nonumber
  &+
  \frac{\rho}{2} \sum_{n=1}^{N}
    \bigl\|\boldsymbol{X}_{(n)} - \mathbf{M}_n + \mathbf{U}_n \bigr\|_F^2.
\end{align}

\subsection{\ac{ADMM} Subproblems}
\label{sec:subproblems}
\ac{ADMM} minimizes \eqref{eq:AL} by alternating exact minimizations over $\mathcal{X}$, each $\mathbf{M}_n$, and dual ascent on each $\mathbf{U}_n$.
Due to the nature of the problem, all three updates can be written in closed form as follows.

\subsubsection{$\mathcal{X}$-Update.}
Minimizing $\mathcal{L}_\rho$ over $\mathcal{X}$ with $\mathbf{M}_n, \mathbf{U}_n$ fixed and the observation constraint \eqref{eq:split_obs} enforced by projection, the objective separates across tensor entries. 
For an unobserved entry $j \in \bar{\Omega}$, the entry $x_j$ appears in $N$ quadratic terms (one per mode unfolding), giving the aggregate problem
\begin{equation}
  \min_{x_j} \; \frac{\rho}{2} \sum_{n=1}^{N}
    \Bigl(x_j - \bigl[\mathbf{M}_n - \mathbf{U}_n\bigr]_j\Bigr)^2,
  \label{eq:x_subproblem}
\end{equation}
whose exact minimizer is the consensus average
\begin{equation}
  x_j^{t+1}
  = \frac{1}{N} \sum_{n=1}^{N}
    \Bigl[\mathrm{Fold}_n\!\left(\mathbf{M}_n^t - \mathbf{U}_n^t\right)\Bigr]_j
  \triangleq v_j^t,
  \label{eq:x_update}
\end{equation}
where the observed entries are fixed, such that $x_j = [\mathcal{T}]_j$ for $j \in \Omega$.

\subsubsection{$\mathbf{M}_n$-Update.}
Next, with $\mathcal{X}^{t+1}$ available the $\mathbf{M}_n$-subproblem is given by
\begin{equation}
  \min_{\mathbf{M}_n} \;
  \lambda \alpha_n \|\mathbf{M}_n\|_*
  + \frac{\rho}{2}
  \bigl\|\hat{\mathbf{X}}_{(n)}^t - \mathbf{M}_n + \mathbf{U}_n^t\bigr\|_F^2.
  \label{eq:M_subproblem}
\end{equation}

Following \cite{Eckstein1992,Boyd2011}, we substitute the plain unfolding
$\boldsymbol{X}_{(n)}^{t+1}$ with the over-relaxed version
\begin{equation}
  \hat{\mathbf{X}}_{(n)}^t
  = \xi \, \boldsymbol{X}_{(n)}^{t+1}
  + (1 - \xi) \, \mathbf{M}_n^t
  \label{eq:overrelax}
\end{equation}
in both the $\mathbf{M}_n$- and $\mathbf{U}_n$-updates, where $\xi \in (1, 2)$ is the over-relaxation factor.  
Setting $\xi = 1$ recovers standard \ac{ADMM}.  
For $\xi \in [1.5, 1.8]$, convergence is provably maintained \cite{Eckstein1992} while iteration counts are reduced by approximately 30--40\% empirically.  
We use $\xi = 1.7$ throughout, following the standard recommendation of \cite{Boyd2011}.

Since it was shown that the proximal operator of the nuclear norm is singular value thresholding \cite{Cai_2010}, the exact closed-form solution is
\begin{equation}
  \mathbf{M}_n^{t+1}
  = \mathrm{SVT}_{\tau_n}\!\left(
      \hat{\mathbf{X}}_{(n)}^t + \mathbf{U}_n^t
    \right),
  \label{eq:M_update}
\end{equation}
where the SVT threshold $\tau_n = \alpha_n \lambda$ is fixed and independent of
$\rho$.  

\subsubsection{$\mathbf{U}_n$-Update.}
In the final step, the dual variable is updated by gradient ascent on the dual function, which yields the closed-form update
\begin{equation}
  \mathbf{U}_n^{t+1}
  = \mathbf{U}_n^t
  + \hat{\mathbf{X}}_{(n)}^t
  - \mathbf{M}_n^{t+1}.
  \label{eq:U_update}
\end{equation}

\subsection{Adaptive Penalty Parameter}
\label{sec:adaptive}

Standard \ac{ADMM} with fixed $\rho$ converges at the $O(1/k)$ rate \cite{Boyd2011}, and the convergence constant depends on how well $\rho$ balances primal and dual feasibility.  
We adopt the adaptive rule of \cite{Boyd2011}, which updates $\rho$ based on the primal and dual residuals computed at each iteration, such that they are defined as
\begin{align}
  r^t &= \sqrt{\sum_{n=1}^{N}
    \bigl\|\boldsymbol{X}_{(n)}^t - \mathbf{M}_n^t\bigr\|_F^2},
  \label{eq:prim_res}\\
  s^t &= \rho^t \sqrt{\sum_{n=1}^{N}
    \bigl\|\mathbf{M}_n^t - \mathbf{M}_n^{t-1}\bigr\|_F^2},
  \label{eq:dual_res}
\end{align}
where the primal residual $r^t$ measures the degree of constraint violation, $i.e.$, how far $\mathcal{X}$ and $\mathbf{M}_n$ are from consensus, while the dual residual $s^t$ measures the rate at which the $\mathbf{M}_n$ variables are changing, $i.e.$, the gradient of the dual function, scaled by $\rho$.

The corresponding update rule for $\rho$ is designed to keep $r^t$ and $s^t$ within a factor of $\mu$ of each other, by increasing $\rho$ when the primal residual is too large, and decreasing it when the dual residual is too large, such that the updated $\rho^{t+1}$ is given by
\begin{equation}
  \rho^{t+1} =
  \begin{cases}
    \min\!\left(\tau \rho^t,\, \rho_{\max}\right)
      & \text{if}\; r^t > \mu\, s^t, \\
    \max\!\left(\rho^t / \tau,\, \rho_{\min}\right)
      & \text{if}\; s^t > \mu\, r^t, \\
    \rho^t & \text{otherwise,}
  \end{cases}
  \label{eq:rho_update}
\end{equation}
with $\mu = 10$, $\tau = 2$, $\rho_{\min} = 0.01$, and $\rho_{\max} = 1000$ in our experiments.  
After each change, the dual variables are rescaled as $\mathbf{U}_n \leftarrow (\rho^t / \rho^{t+1})\,\mathbf{U}_n$ to preserve the value of the scaled augmented Lagrangian \eqref{eq:AL} across the parameter change \cite{Boyd2011}.

\paragraph{Data-driven initialization.}
The initial penalty $\rho^0$ is set to balance the scale of the SVT threshold against the singular value spectrum of the initial estimate.
Specifically, let $\bar{\sigma}$ denote the mean of the non-zero singular values of $\boldsymbol{X}_{(n)}^0$ averaged across modes. 
Thus, the initial penalty is defined as
  \begin{equation}
      \rho^0 = \frac{\bar{\sigma} }{N\cdot\lambda},
  \end{equation}
so that at iteration 1 the SVT threshold $\tau_n = \alpha_n \lambda$ operates near the centre of the singular value spectrum, making the first $\mathbf{M}_n$-update maximally informative.


The complete procedure is summarized in Algorithm \ref{alg:snn_admm}.

\begin{algorithm}[t]
\caption{ADMM-LRTC: Low-Rank Tensor Completion via Adaptive Over-Relaxed \ac{ADMM}}
\label{alg:snn_admm}
\begin{algorithmic}[1] 
\Statex \textbf{Input:} Partially observed tensor $\mathcal{T}$, index set $\Omega$,
\Statex constants $\lambda$, $\alpha_n$, $\xi$, $T_{\max}$
\Statex \hspace{-4.4ex} \hrulefill \vspace{-0.3ex}

\State $\mathcal{X}^0 \leftarrow \mathcal{P}_\Omega(\mathcal{T}) + \bar{o}\,\mathcal{P}_{\bar\Omega}(\mathbf{1})$
\State $\mathbf{M}_n^0 \leftarrow \mathrm{SVT}_{\alpha_n\lambda}(\mathcal{X}^0_{(n)})$, $\mathbf{U}_n^0 \leftarrow \mathbf{0}$ \text{ for } $n=1,\ldots,N$
\State $\rho^0 \leftarrow \bar\sigma\,\bar\alpha / \lambda$ via \eqref{eq:rho_init}

\For{until convergence, or $t = 1, 2, \ldots, T_{\max}$}
  \State Compute consensus $v_j^t$ via \eqref{eq:x_update}
  \State $\mathcal{X}^{t+1}_{j} \leftarrow v_j^t \quad \forall j \in \bar\Omega$
  \State $\mathcal{X}^{t+1}_{j} \leftarrow [\mathcal{T}]_j \quad \forall j \in \Omega$
  
  \For{$n = 1, \ldots, N$}
    \State $\hat{\mathbf{X}}_{(n)}^t \leftarrow \xi\,\boldsymbol{X}_{(n)}^{t+1} + (1-\xi)\,\mathbf{M}_n^t$
    \State $\mathbf{M}_n^{t+1} \leftarrow \mathrm{SVT}_{\alpha_n\lambda}(\hat{\mathbf{X}}_{(n)}^t + \mathbf{U}_n^t)$
    \State $\mathbf{U}_n^{t+1} \leftarrow \mathbf{U}_n^t + \hat{\mathbf{X}}_{(n)}^t - \mathbf{M}_n^{t+1}$
  \EndFor
  
  \State Compute $r^t$, $s^t$ via \eqref{eq:prim_res}, \eqref{eq:dual_res}
  \State Update $\rho^{t+1}$ via \eqref{eq:rho_update}; rescale $\mathbf{U}_n^{t+1}$

\EndFor

\State \Return $\mathcal{X}^\star = \mathrm{clip}_{[0,255]}(\mathcal{X}^{t+1})$
\end{algorithmic}
\end{algorithm}

\section{Numerical Results}
\label{sec:results}
\begin{figure}
    \includegraphics[width=\columnwidth]{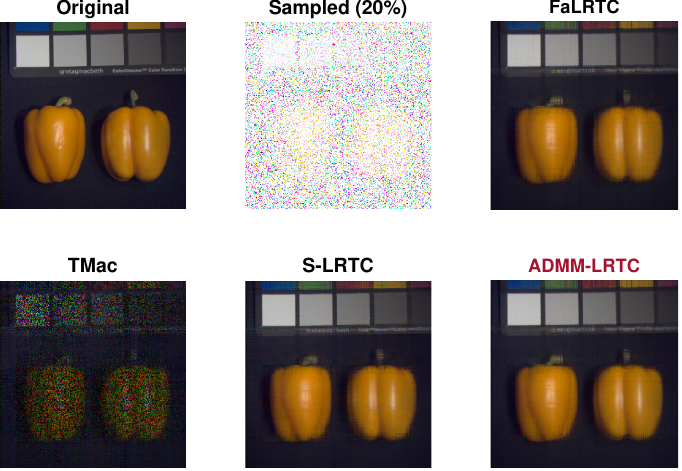}
    \vspace{-4ex}
    \caption{Illustration of the completed test image for the proposed method, compared to the original image, the sampled image and the \ac{SotA} methods reconstructed images for an observation ratio of $20\%$.}
    \label{fig:image}
\end{figure}

\vspace{-1ex}
In this final section we compare the proposed method to the \ac{NN} minimization-based \ac{SotA} technique (FaLRTC) described in \cite{Liu_2013} and a more recent technique that uses a mixture of \ac{NN} minimization and matrix factorization (S-LRTC), as described in \cite{Gao2018}.
For comparison, an arbitrary image\footnotemark presented in Figure \ref{fig:image} is chosen, which is of size $256\times 256$ and contains RGB color channels represented as a third dimension, leading to a tensor of size $256\times 256\times 3$. 
The chosen performance metric for comparison, conventionally used in the \ac{SotA}, is the \ac{NMSE}, defined as
\vspace{-1ex}
\begin{equation}
\mathrm{NMSE} = \frac{\|\widehat{\mathcal{X}} - \mathcal{X}_{\mathrm{true}}\|_F^2}{\|\mathcal{X}_{\mathrm{true}}\|_F^2}.
\label{eq:NMSE}
\end{equation}
\vspace{-2ex}

\begin{figure}[H]
    \centering
    \includegraphics[width=\columnwidth]{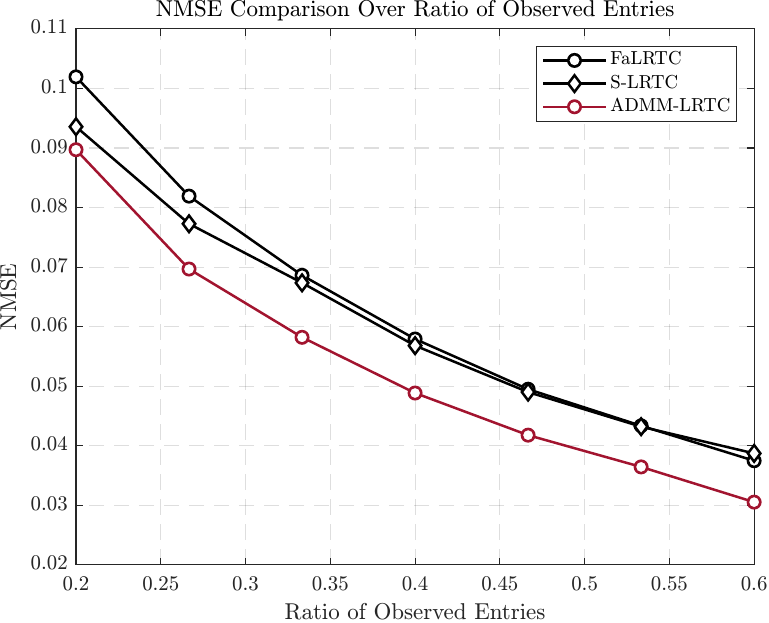}
    \vspace{-4ex}
    \caption{\ac{NMSE} comparison of the \ac{SotA} and the proposed method, with a varying ratio of observed entries in $\boldsymbol{O}$.}
    \label{fig:NMSE}
\end{figure}

\begin{figure}[H]
    \includegraphics[width=\columnwidth]{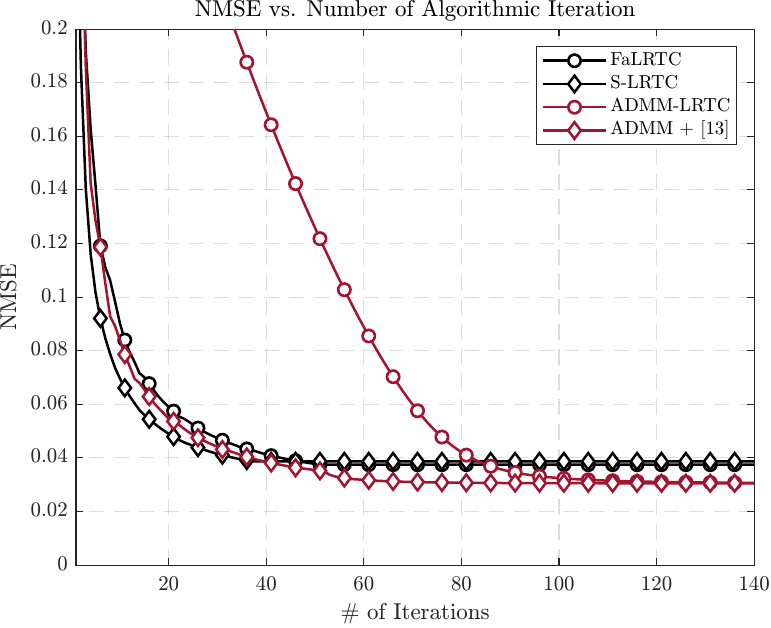}
    \vspace{-4ex}
    \caption{\ac{NMSE} convergence comparison of the \ac{SotA} and the proposed method, with a $60\%$ observation ratio of $\boldsymbol{O}$.}
    \label{fig:conv}
\end{figure}

To compare the \ac{SotA} methods with the proposed method under the same conditions, identical simulation setups are chosen, where the ratio of observed values in the tensor $\boldsymbol{O}$ varies from $20\%$ to $60\%$.
The \ac{NMSE} results over varying observation ratios are shown in Figure \ref{fig:NMSE}, while the convergence behavior for a fixed observation ratio of $60\%$ is shown in Figure \ref{fig:conv}.

In Figure \ref{fig:NMSE}, it can be observed that the proposed method clearly outperforms the \ac{SotA} methods across all observation ratios.
To elaborate, to reach an \ac{NMSE} of $0.05$, the proposed method requires only approximately $39\%$ of the observed entries, while the \ac{SotA} methods require at least $46\%$ of the observed entries to reach the same \ac{NMSE} level, which demonstrates the superior performance of the proposed method in terms of reconstruction accuracy.

Next in Figure \ref{fig:conv}, the convergence behavior of the proposed method is compared to the \ac{SotA} methods for a fixed observation ratio of $60\%$.
It can be observed that the proposed method converges significantly slower than the \ac{SotA} methods, however, it achieves a much lower \ac{NMSE} level at convergence.

\vspace{-3ex}

\footnotetext{The chosen test image is available at:\\ https://github.com/shangqigao/TensorCompletion}
To further investigate the convergence behavior of the proposed method and improve its convergence speed, an additional simulation is conducted, where the proposed method is initialized with the FaLRTC \ac{SotA} method reconstructed tensor, instead of a random initialization.
As also illustrated in Figure \ref{fig:conv}, the proposed method with the FaLRTC initialization converges significantly faster than the proposed method with random initialization, while achieving the same \ac{NMSE} level at convergence, which demonstrates that the convergence speed of the proposed method can be improved by using a better initialization.

Finally, for the sake of completeness, Figure \ref{fig:image} shows the visual result of the \ac{SotA} methods compared to the proposed method for a fixed observation ratio of $20\%$, along with the original image and the sampled image that is used as the input for algorithms.
Along the proposed method and the \ac{SotA} methods shown before, the TMac method is also included in the visual comparison \cite{xu2013parallel}, which is a matrix factorization-based method, but was neglected in the \ac{NMSE} comparison due to its significantly worse performance compared to the other methods.

\vspace{-2ex}

\section{Conclusion}
\label{sec:conclusions}
\vspace{-1ex}

We proposed a novel \acf{TC} approach in which \acf{ADMM} in addition to an adaptive penalty, as well as over-relaxation is used to solve the \ac{NN} minimization problem.
The resulting ADMM method is shown to significantly outperform the \ac{SotA} in terms of \ac{NMSE} performance, across all observation ratios, albeit at the cost of a slower convergence speed.
Additionally, it is shown that the convergence speed of the proposed method can be improved by using a better initialization, such as the reconstructed tensor from the \ac{SotA} method, without affecting the \ac{NMSE} performance at convergence.
Future work will focus on the improvement the proposed method, both in terms of \ac{NMSE} performance and convergence.
\bibliographystyle{IEEEtran}
\bibliography{IEEEabrv,ref_paper.bib}

\end{document}